\pdfoutput=1

\documentclass[11pt]{article}

\usepackage[final]{acl}

\usepackage{times}
\usepackage{graphicx}
\usepackage{latexsym}
\usepackage{titlesec}
\usepackage{array}
\usepackage{multicol}
\usepackage{booktabs}
\usepackage{multirow}
\usepackage{amssymb}
\usepackage{rotating}
\usepackage{longtable}
\usepackage{tabularx}
\usepackage{pdflscape}
\usepackage{enumitem}

\setlist[enumerate]{itemsep=2pt}
\titleformat*{\subsection}{\fontsize{11.5pt}{15pt}\selectfont\bfseries}

\usepackage[T1]{fontenc}

\usepackage[utf8]{inputenc}

\usepackage{microtype}
\usepackage{amsmath}

\usepackage{makecell}

\title{Towards a Neural Era in Dialogue Management for Collaboration: A Literature Survey}

\author{Amogh Mannekote \\
  University of Florida \\
  \texttt{amogh.mannekote@ufl.edu}}

\newcommand{\archdiagram}[4][1]{
    \begin{figure*}[h!]
    \centering
    \resizebox{#1\textwidth}{!}{
        \includegraphics{figures/#2/#3.png}
    }
    \caption{#4}
    \label{fig:#3}
    \end{figure*}
}

\newcommand{\archdiagramhalf}[4][1]{
    \begin{figure}[h!]
    \centering
    \resizebox{#1\textwidth}{!}{
        \includegraphics{figures/#2/#3.png}
    }
    \caption{#4}
    \label{fig:#3}
    \end{figure}
}

\newcommand{\rqone}{What dialogue management approaches have been utilized to develop collaborative dialogue agents?}
\newcommand{\rqtwo}{What techniques have been applied to adapt neural network based dialogue management approaches to collaborative settings?}
\newcommand{\numworks}{17 }

\begin{document}
\maketitle

\begin{abstract}
Dialogue-based human-AI collaboration can revolutionize collaborative problem-solving, creative exploration, and social support. To realize this goal, the development of automated agents proficient in skills such as negotiating, following instructions, establishing common ground, and progressing shared tasks is essential. This survey begins by reviewing the evolution of dialogue management paradigms in collaborative dialogue systems, from traditional handcrafted and information-state based methods to AI planning-inspired approaches. It then shifts focus to contemporary data-driven dialogue management techniques, which seek to transfer deep learning successes from form-filling and open-domain settings to collaborative contexts. The paper proceeds to analyze a selected set of recent works that apply neural approaches to collaborative dialogue management, spotlighting prevailing trends in the field. This survey hopes to provide foundational background for future advancements in collaborative dialogue management, particularly as the dialogue systems community continues to embrace the potential of large language models.

\end{abstract}

\section{Introduction}
Human collaboration through dialogue is a fundamental aspect of our everyday lives, as we often engage in conversations to work together, solve problems, and achieve shared goals \cite{streeck2011embodied}. However, effective collaboration through dialogue is a complex process, as it requires the participants to take turns, negotiate meaning \cite{Clark1991GroundingIC}, coordinate actions \cite{clark1986referring}, and manage conflicts \cite{traum_multi-party_2008}, all while maintaining a shared understanding of the task at hand. These factors contribute to the complexity of building an automated dialogue system that can act as a collaborative partner to one or more humans. 

It has been a longstanding goal in dialogue systems research to develop domain-specific systems, general-purpose frameworks, and algorithms to build automated agents that can act as collaborative partners to human users \cite{rich_collagen_2001, bohus_ravenclaw_2009, 10.1145/544862.544923, galescu_cogent_2018}. However, most approaches that predate the use of neural network models require some form of manual labor to extend their functionality to a new domain. Recently, the advent of deep learning based approaches ushered in a new era of data-driven methods for carrying out various tasks within a dialogue system, including natural language understanding \cite{louvan2020recent, shah2019robust}, dialogue management \cite{mrkvsic2016neural, ren2018towards}, and natural language generation \cite{wen2015semantically, santhanam2019survey}. In fact, a long line of work from the mid-2010s has produced success stories in the realms of form-filling agents (e.g., used to carry out tasks such as booking a restaurant table on behalf of the user) and open-domain agents (also called chatbots, which engage in open-ended conversation with the user with no particular end-goal). Nevertheless, compared to form-filling and open-domain agents, there has been a paucity in both theoretical advances and practical implementations that use neural-network based approaches to build agents for collaborative dialogue.

In this survey, I first explore traditional dialogue management paradigms used for over three decades in developing automated agents for collaborative dialogue. Dialogue management paradigms are abstract computational models offering guidelines for processing inputs, guiding conversations, and maintaining context. Concrete data structures and algorithms translate these paradigms into functioning systems. I provide a concise description of each paradigm, emphasizing its applications in collaborative contexts, in Section \ref{sec:rq1}.

The survey then narrows its focus to neural network-based approaches for collaborative dialogue management. Since the mid-2010s, various methodologies have sought to combine traditional paradigm strengths with the representational flexibility of neural networks. This integration involves modifying standard end-to-end neural architectures, typically used for form-filling and open-domain settings, which is explored in Section \ref{sec:rq2}.

This literature survey systematically explores these themes through the following two research questions (RQs):
\begin{itemize}
\item \textbf{RQ 1:} \rqone
\item \textbf{RQ 2:} \rqtwo
\end{itemize}

The remainder of this paper is organized as follows. Section \ref{sec:background} provides background on collaborative dialogue systems and the dialogue management problem. In Section \ref{sec:methods}, I list the research questions and discuss the literature search process. Section \ref{sec:rq1} examines various design paradigms for building collaborative dialogue managers. Before answering RQ 2, Section \ref{sec:challenges-neural} outlines challenges faced by neural approaches to collaborative dialogue management. In Section \ref{sec:rq2}, I focus on adaptations made to neural-network-based dialogue management approaches for collaborative settings. Section \ref{sec:discussion} discusses survey findings and future research directions, while \ref{sec:conclusion} concludes the survey with a summary.

\section{Background} \label{sec:background}
This section begins by providing a working definition of collaborative dialogue systems. Then, it highlights the role of the dialogue manager within a dialogue system, which is the module responsible for guiding the conversation.

\subsection{Defining Collaborative Dialogue Systems}

Collaborative dialogue systems are designed to collaborate with human users to achieve one or more shared goals, such as co-creating an artifact or navigating a physical environment \cite{streeck2011embodied}. These systems must support functions such as negotiating shared goals and developing a mental model of the human user. To better understand collaborative dialogue systems, it is helpful to first introduce two other types of dialogue agents: form-filling assistants and chatbots, and then contrast them with collaborative dialogue agents.

Form-filling assistants, also known as task-oriented dialogue systems \cite{williams2014dialog}, perform tasks on the user's behalf, such as reserving a table or booking a flight. However, they do not possess their own goals or agency. Chatbots, on the other hand, engage users in open-ended conversations without specific goals, making them non-collaborative as well \cite{huang2020challenges}. In contrast to these two types of dialogue agents, collaborative dialogue agents play a variety of roles that involve working together with the human user.

Some examples of roles that collaborative dialogue agents may assume include that of an information giver (providing step-by-step instructions to the human user to follow), a collaborative partner (planning and working together with the human to complete a task), a tutor (helping the human user learn a concept or a skill), and even an opponent (in a negotiation setting or within the context of a game). These agents have their own goals that may be slightly or even greatly different from that of the human user (e.g., an agent designed to bargain with a human).

\subsection{Role of the Dialogue Manager}
Computational approaches to dialogue serves two distinct purposes: dialogue modeling and dialogue management \cite{edda_computational_2017}. Dialogue modeling focuses on understanding and explaining conversation dynamics without actively participating. In contrast, dialogue management involves deciding what to say next based on the dialogue context, thereby enabling the agent to actively participate in the conversation. This survey is specifically concerned with dialogue management.

The dialogue manager guides the conversation within a dialogue system. Conceptually, it can be divided into representation and decision-making. Figure \ref{fig:pipeline2} shows a standard pipeline architecture of a dialogue system. Note that while most dialogue systems can be factored into these components, it may not always be the case. In this pipeline architecture, dialogue management is carried out jointly by the state tracker and the dialogue policy. The representation, or dialogue state, provides an internal summary of the current dialogue status, which needs to be continuously updated throughout the conversation. The decision-making aspect, or dialogue policy, determines the most suitable action based on the dialogue state and other relevant factors, such as the state of a shared task. In some approaches, such as rudimentary script-based approaches, these two functions may be inseparably coupled into a single functional module.

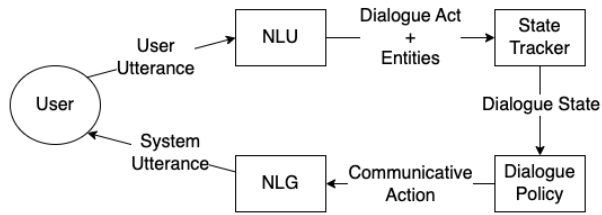
\archdiagramhalf[0.5]{rq1}{pipeline2}{A pipeline-based architecture of a dialogue system, comprising modules for natural language understanding (NLU), state tracking, dialogue policy, and natural language generation (NLG). The state tracking and dialogue policy are jointly responsible for carrying out the function of dialogue management.}

\section{Methods}
\label{sec:methods}

This section outlines the search methodology, including the inclusion and exclusion criteria employed to select relevant works for the literature survey. Then, it details the approach taken to identify the modeling themes for answering RQ 2.

\subsection{Search Strategies}

The search strategy for RQ 1 and RQ 2 were conducted independently.

\subsubsection{Search Strategy for RQ 1}
To address RQ 1, the search began with a curated ``seed'' collection of influential works presenting foundational dialogue management paradigms for collaborative dialogue. Additional works were identified through the reference lists of these highly cited publications (snowballing). Specifically, the inclusion and exclusion criteria for this RQ were as follows:
\begin{itemize}
    \item \textbf{IC1:} The work should present either a novel paradigm of dialogue management or a domain-independent framework that can be leveraged to develop a wide range of domain-specific dialogue systems.
    \item \textbf{IC2: } A sizeable number of practical dialogue system implementations should be implemented using the paradigm.
    \item \textbf{IC3: } The dialogue management paradigm should have been used to develop one or more of the following types of collaborative agents: collaborative problem-solving, negotiation, or tutoring.
    \item \textbf{EC1: } The system or framework should not specific to one particular domain, in which case it cannot be considered to be a ``paradigm.''
\end{itemize}
From reviewing these combined set of works, I identified five distinct paradigms of collaborative dialogue management: script-based, plan-based, information-state based, those based on \citet{10.1145/544862.544923}'s collaborative problem solving (CPS) framework, and neural-network based approaches. 

\subsubsection{Search Strategy for RQ 2}
For RQ 2, a separate and distinct search was conducted with the goal of discovering neural network-based approaches to developing collaborative dialogue managers. A targeted search was performed using terms such as "collaborative dialogue systems," "collaborative dialogue state," and "collaborative dialogue" across Association for Computational Linguistics (ACL) and search engines such as Zeta Alpha\footnote{\href{https://zeta-alpha.com}{zeta-alpha.com}}, Semantic Scholar\footnote{\href{https://semanticscholar.org}{semanticscholar.org}}, and Google Scholar\footnote{\href{https://scholar.google.com/}{scholar.google.com}}. The inclusion and exclusion criteria for selecting papers were applied specifically for RQ 2, as follows:
\begin{itemize}
    \item \textbf{IC1:} Inclusion of neural network-based approaches to developing collaborative dialogue managers.
    \item \textbf{IC2:} Consideration of works primarily published post-2016, reflecting the emergence of neural network approaches during this period.
    \item \textbf{IC3: } Negotiation dialogue systems; although commonly characterized as non-collaborative dialogue, these systems were included as negotiation frequently features in collaborative dialogue during planning and shared goal setting.
    \item \textbf{EC1:} Exclusion of works that lack a data-driven, neural network-based dialogue manager implementation.
    \item \textbf{EC2:} Exclusion of studies focusing on multi-party collaborative dialogue systems, with the scope limited to two-party systems.
    \end{itemize}

The final set of papers was derived from a combination of this search strategy for RQ 2 and additional publications discovered through the citation networks of these articles that also fit these criteria.




\section{RQ 1: \rqone}
\label{sec:rq1}

In this section, I describe script-based, plan-based, information-state, and neural-network based paradigms to collaborative dialogue management. In addition, I also describe the Collaborative Problem Solving (CPS) framework, which was designed specially with collaborative dialogue management in mind. Except for the CPS framework, I begin with a general description of each paradigm and then touch upon how they have been used specifically for collaborative settings.

\subsection{Script-based Approaches} \label{sec:structure}
Script-based approaches to dialogue management are based on a predetermined, ``ideal'' flow of dialogue and are handcrafted to each specific domain \cite{edda_computational_2017}. Specifically, it involves specifying a handcrafted  ''script'' that both the dialogue system and the user must follow. A script usually consists of fixed patterns and pre-written responses, which can be matched to user input through pattern matching, keyword extraction, or rule-based methods. For instance, a script can encode an adjacency-pair rule stating that a question needs to be followed by an answer \cite{dahlback-jonsson-1989-empirical}. More complex scripts are also equipped to handle a rudimentary level of deviations, such as clarification questions in the dialogue. Examples of popular, early implementations of script-based dialogue systems include ELIZA \cite{weizenbaum1966eliza}, LINLIN \cite{dahlback1999architechture}, and RailTel \cite{bennacef1996dialog}. 


Within collaborative contexts, one popular type of dialogue systems where script-based dialogue management is extensively employed is dialogue-based intelligent tutoring systems (ITSs) \cite{webb_developing_1999, Strijbos2004TheEO}. ITSs are adaptive educational systems that provide personalized instruction and feedback to learners, employing techniques from artificial intelligence, cognitive psychology, and education research to optimize the learning process \cite{wollny_are_2021}. Many ITSs such as CIRCSIM-Tutor \cite{evens_circsim-tutor_1997}, AutoTutor \cite{graesser_autotutor_2004} and Why2-Atlas \cite{vanlehn_architecture_2002} make use of dialogue-grammar based (also called script-based) dialogue managers. These grammars (or scripts) are designed in such a way so as to help the user achieve a goal.

For example, the grammar in Why2-Atlas \cite{vanlehn_architecture_2002} enforces the student to provide a comprehensive explanation of a physics phenomenon. It works by mapping utterances to an explanation or a misconception using pattern matching techniques. Once the agent believes it has identified the right explanation or misconception, it then responds with a pre-defined response corresponding to it. The dialogue manager continues to cycle through the list of all pieces of the unmentioned explanations until it is satisfied that the student understands the concept.

While handcrafted, script-based approaches can be useful for creating simple, controlled dialogue systems with predictable conversation flows, they tend to lack flexibility and adaptability. As a result, they might be limited in their ability to handle diverse, complex, or unanticipated user inputs, which can limit their effectiveness in more dynamic and interactive scenarios. Despite these limitations, handcrafted dialogue management has laid the foundation for more advanced techniques, and it continues to be employed in specific contexts where a well-defined, structured conversation is required.

\subsection{Plan-Based Approaches}
\label{sec:rq1-planning}
Humans naturally ascribe intentions and goals to their conversational counterparts, expecting rational behavior in line with these objectives. Plan-based methodologies for dialogue management encapsulate this collaborative mindset \cite{papaioannou2018human}. One notable implication of this viewpoint is that natural language expressions are perceived as actions (dialogue acts \cite{searle_1969}) undertaken by rational agents to achieve goals \cite{allen1983recognizing, pollack1992uses}. Several systems also make use of plan recognition to deduce user's objective based on their actions. The goal-oriented nature of this paradigm offers a degree of coherence and consistency in a dialogue agent's behavior that is challenging to attain through alternative dialogue management paradigms. As such, this approach has been adopted in high-stakes application domains such as healthcare \cite{vannaprathip_intelligent_2022, kane_flexible_2022} and education \cite{freedman_plan-based_2000}.

Plan-based dialogue management originates from AI planning \cite{ghallab2004automated}, which can be described as a search problem aimed at identifying a sequence of \textit{actions} that, upon execution, accomplishes a goal. At any given moment, the agent resides in one of several potential \textit{states}. Actions transition the system from one state to another. Each action possesses a set of \textit{preconditions}, which must be satisfied for the action to be deemed eligible by the planning algorithm. The outcome of performing an action is defined as a series of \textit{effects} on the state. Ultimately, a planning problem is initiated with both an initial and a final goal state.

Over the past three decades, numerous practical implementations of plan-based dialogue management in collaborative dialogues have been proposed. Two notable implementations of plan-based dialogue managers in collaborative problem-solving settings are the TRAINS \cite{allen1995trains} and TRIPS \cite{allen2005two} systems by James Allen's group. More recent works integrate mixed-initiative capabilities, such as MAPGEN \cite{ai2004mapgen}, which implemented a plan-based system for ground operations of a NASA Mars Rover and SIADEX \cite{de2005siadex}, which was deployed to help coordinate forest-fire operations. Various domain-independent frameworks such as COLLAGEN \cite{rich_collagen_2001}, PASSAT \cite{myers2002passat}, and RavenClaw \cite{bohus_ravenclaw_2009} have been developed to allow domain-specific agents to be built.

\subsection{Information-State Based Approaches} \label{sec:informational}

In the information-state based approach to dialogue management, as proposed by \citet{larsson2000information}, the dialogue state is defined by a collection of beliefs, desires, and intentions (BDI) \cite{bdi}. As the dialogue unfolds, each utterance leads to an update in the dialogue state. The precise nature of these updates relies on a series of update rules and strategies. This approach allows for a high level of adaptability due to its minimal constraints on the dialogue manager, making it suitable for various dialogue situations, including collaboration.

When examining the information-state approach more closely, it can be characterized by three primary components: informational components, dialogue acts, and update rules \cite{larsson2000information}.

\begin{enumerate}
\item \textbf{Informational Components.} In collaborative settings, information states are categorized into private aspects (individual agent's beliefs, desires, and intentions) and shared aspects (mutually agreed-upon beliefs, desires, and intentions). These components can be either static, such as agent persona and background knowledge, or dynamic, such as evolving goals and dialogue history.

\item \textbf{Formal Representations.} Various representations may be used for informational components, affecting their accessibility, ease of updating, and overall efficiency. These representations can include basic data structures, composite data types, logical forms, or neural embeddings.

\item \textbf{Dialogue Moves.} Dialogue moves represent the "illocutionary force" of an utterance as defined by \citet{searle_1969}. They are identified using domain-specific coding schemes.

\item \textbf{Update Rules.} Drawing inspiration from plan-based dialogue management (see Section \ref{sec:rq1-planning}), update rules consist of preconditions and effects \cite{traum_dialogue_moves}. A rule can only be applied if its preconditions are met; once they are, the effects are used to update the information state.

\item \textbf{Control Strategy.} Control strategies dictate the selection and sequence of applied update rules. These strategies can involve: 1) applying the first rule that matches, 2) sequentially applying all matching rules, 3) employing a probabilistic rule for selection, or 4) allowing the user to make the choice.

\end{enumerate}



TrindiKit is a domain-independent framework for collaborative dialogue management \cite{larsson2000information}. Basilica is another example of a domain-independent dialogue management framework proposed by \citet{kumar_architecture_2011}. It uses a unified architecture for authoring tutoring and collaborative learning agents. Basilica works on a decentralized network of object-oriented modules, each one programmed and functioning independently through common interfaces. Basilica has been used to develop several collaborative learning agents, including CycleTalk tutor \cite{roman_rmm_2020} and TuTalk \cite{jordan_tutalk}. To share information and events across modules, Basilica uses a shared, global state. This state corresponds to the information state of the agent. Modules can perform a wide variety of functions, including recognition of student affect, monitoring of task progress, and dialogue management.


\archdiagram[0.70]{rq1}{cps}{Collaborative problem solving model \cite{10.1145/544862.544923}}

\subsection{Collaborative-Problem Solving (CPS) Framework}
\citet{10.1145/544862.544923} introduced the domain-independent CPS framework for dialogue management collaborative agents. The CPS model is designed from ground up to handle more complex tasks than state-based or script-based models (such as those in Sections \ref{sec:structure} and \ref{sec:informational}). The model views a collaborative dialogue as operating over three levels of interaction (see Figure \ref{fig:cps}):
\begin{enumerate}
    \item \textbf{Problem Solving Level:} This level describes how a single agent solves problems. It includes processes such as planning, decision-making, and execution.

    \item \textbf{Interaction Level:} This level involves the interaction between two or more agents. It includes processes such as communication, negotiation, and coordination.

    \item \textbf{Coordination Level:} This level involves the coordination of multiple interactions between agents to achieve a common goal. It includes processes such as resource allocation, task assignment, and conflict resolution.
\end{enumerate}

The CPS model involves multiple agents working together to solve a problem, with each agent contributing its own expertise and knowledge to the process. While the authors are primarily focused on human-machine collaboration, they believe that the model will equally well apply to interactions between sophisticated software agents that need to coordinate their activities.

Based on the CPS model, \citet{galescu_cogent_2018} introduce the Cogent dialogue shell as a practical, extensible, domain-independent ``shell'' on top of which domain-specific agents can be built. It achieves this by providing many of the domain-independent capabilities out-of-the-box, including general-purpose natural language parsing and collaborative dialogue management capabilities.


\subsection{Neural-Network Based Approaches}
\label{sec:rq1-e2e}
In their most general form, neural-network based dialogue managers use sequence-to-sequence models \cite{sutskever2014sequence} to generate the dialogue system's next response directly from the natural language dialogue history. The advantage of building a dialogue manager this way is that it circumvents labor-intensive handcrafting of dialogue management strategies by learning it from a corpus of \textit{training} dialogues. These corpora are typically collected from human-human interactions in the same or similar domain. This paradigm of training a dialogue agent was originally popularized in the context of open-domain chatbots \cite{shum2018eliza, wolf2019transfertransfoatransfer, serban2016building}. However, several works soon followed it up by extending the paradigm to task-oriented dialogue systems that performed slot-filling \cite{bordes2016learning, wen_network-based_2017}.






\subsubsection{Challenges in Neural Collaborative Dialogue Management} \label{sec:challenges-neural}
Before diving deep into the second research question, I enumerate some of the key challenges associated with neural network-based approaches to building collaborative dialogue systems as background.



\begin{enumerate}

\item \textbf{Lack of interpretability and explainability:} The absence of explicit internal structure and intermediate representations in end-to-end models can make it difficult to interpret and understand the decision-making process, which is particularly important in strategic and collaborative settings where trust and explainability play a critical role.

\item \textbf{Difficulty in grounding to real-world entities:} Collaborative dialogue often takes place within the context of a common artifact such as a co-observed image \cite{shinagawa_interactive_2020, kim-etal-2019-codraw} or an environment \cite{roman_rmm_2020, chi_just_2020}. However, vanilla neural dialogue models\footnote{The term "vanilla" is used to describe unmodified or basic versions of architectures, without any special enhancements or adaptations. It refers to the fundamental, original form of a model, serving as a foundation for more advanced or customized variations.} often struggle with grounding, as they lack explicit mechanisms to represent or reason about real-world entities, leading to contextually inappropriate or inconsistent responses. 

\item \textbf{Large amounts of training data required:} These models often need extensive training data to achieve good performance, which may not be readily available for specialized or niche domains \cite{sutskever2014sequence, li_deep_2016, serban16hierarchical}.

\item \textbf{Hallucinations and challenges in generating accurate responses:} In the context of generative language models, a hallucination refers to a situation where the AI model generates output that is not grounded in the input data, factual information, or the intended context \cite{ji2023survey, cao2022hallucinated}. This can happen due to several reasons, such as overfitting, biases in the training data, or insufficient constraints on the generative process. For example, if a language model is asked to generate an answer to a historical question and it produces a response with incorrect dates or events, that would be considered a hallucination. Hallucinations are, therefore, a challenge to be dealt with when using neural models for generating dialogue responses, decreasing the overall quality and reliability of the conversation.

\item \textbf{Difficulty incorporating external knowledge and adapting to changes:} Such models have trouble incorporating external knowledge or adapting to changes in the environment during a conversation, which is a common requirement in strategic and collaborative applications \cite{sutskever2014sequence, weston-etal-2018-retrieve}.


\end{enumerate}

\section{RQ 2: \rqtwo}
\label{sec:rq2}
To systematically analyze the \numworks selected works for addressing RQ 2, I employed a three-step approach to investigate each work, extract relevant architectural modifications, and classify them into overarching themes. The process unfolded as follows:
\begin{enumerate}
    \item \textbf{In-depth examination:} I analyzed each work, delving into the proposed architectural alterations and evaluating their significance for neural dialogue management.
    
    \item \textbf{Extraction and preliminary grouping:} While assessing each work, I extracted and recorded all architectural changes, observing their distinct attributes and qualities. This process led to an initial collection of categories that encompassed the diverse facets of the modifications.
    
    \item \textbf{Thematic integration:} After completing the extraction and preliminary categorization, I executed an extensive synthesis of the findings. Through a recursive process of analysis and comparison, I discerned connections and patterns among the various categories. 
\end{enumerate}

\onecolumn
\begin{landscape}
  \centering
  \renewcommand{\arraystretch}{1.5} 
  
  \begin{longtable}{>{\raggedright}p{4.2cm}>{\raggedright}p{2.7cm}>{\raggedright}p{3.8cm}cccccc}
    \caption{Overview of different papers and their corresponding modeling themes. Columns with boolean values are marked with a check symbol for true and left empty for false. The columns represent the following: (1) Paper, (2) Participant Roles (* indicates the participant being modeled), (3) Incorporating Shared Artifacts or Environments, (4) Decoupling Semantics and Surface Realization, (5) Graph-Based Representations of Dialogue State, (6) Incorporating Domain-Specific or Expert Knowledge, (7) Theory-of-Mind Modeling, (8) Modality Encoding/Fusion}\label{tab:overview}\\
    \toprule
    \textbf{Paper} & \textbf{Participant Roles} & \textbf{Modalities or Task} & \textbf{Decoupling} & \textbf{Graph} & \textbf{Domain} & \textbf{ToM} & \textbf{Modality} \\
    \midrule
    \endfirsthead
    \caption{Overview of different papers and their corresponding features (Continued)}\\
    \toprule
    \textbf{Paper} & \textbf{Participant Roles} & \textbf{Modalities or Task} & \textbf{Decoupling} & \textbf{Graph} & \textbf{Domain} & \textbf{ToM} & \textbf{Modality} \\
    \midrule
    \endhead
    \hline
    \multicolumn{3}{r}{Continued on next page}
    \endfoot
    \hline
    \endlastfoot
    
    
    \citet{bara_mindcraft_2021} & Player 1*,\newline Player 2* & theory-of-mind questionnaires,\newline Minecraft 3D game state &  &  &  & \checkmark &  \\
    
    \citet{chi_just_2020} & Navigator*,\newline Instructor & 3D room environment &  & \checkmark &  &  & \checkmark &  \\



    \citet{de_vries_talk_2018} & Guide*,\newline Tourist & Map of a City &  & \checkmark &  &  &  &   \\

    \citet{he_learning_2017} & Friend 1*,\newline Friend 2* & Knowledge Bases of Friends &  & \checkmark  &  &  &  &   \\

    \citet{he-etal-2018-decoupling} & Buyer,\newline Seller & Negotiation Scenario + Item Details & \checkmark &  &  &  &  &  \\

    \citet{jayannavar-etal-2020-learning} & Builder*,\newline Architect & Blocks World &  &  &  &  & \checkmark & \\


    \citet{li_end--end_2019} & User*,\newline Attacker & Private Profile Information & \checkmark &  & \checkmark &  &  &   \\


    \citet{kim-etal-2019-codraw} & Teller*,\newline Drawer & Drawing Canvas with Objects &  &  &  &  & \checkmark &  \\

    \citet{narayan-chen_towards_2017} & Builder*,\newline Architect & 2D Blocks World &  &  &  &  & \checkmark &  \\

    \citet{narayan-chen_collaborative_2019} & Builder*,\newline Architect & 3D Minecraft World &  &  &  &  & \checkmark &  \\


    \citet{qiu_towards_2022} & Player*,\newline Other Players* & Game State (including Persona Descriptions) &  & \checkmark & \checkmark & \checkmark &  &  \\


    \citet{santhanam_learning_2020} & Persuader*,\newline Persuadee & Charity Donation & \checkmark &  &  &  &  &   \\

    \citet{shi_learning_2022} & Architect,\newline Builder* & 3D Minecraft World &  &  &  &  & \checkmark &    \\

    \citet{Yarats2017HierarchicalTG} & Negotiator 1*,\newline Negotiator 2* & Private Negotiation Goals + Item Pool & \checkmark &  &  &  &  &   \\

    \citet{zhou-etal-2019-dynamic} & Buyer,\newline Seller,\newline Coach* & Product Listing &  &  & \checkmark &  &  & \\

    \citet{zhou_augmenting_2020} & Negotiatior 1*,\newline Negotiator 2* & Product Listing (for CraigslistBargain) &  &  & \checkmark &  &  &   \\

    \citet{zhou_ai_2022} & DM*,\newline Players  & Game Background Story + Game State  &  &  &  & \checkmark &  &  \\
    
  \end{longtable}

\end{landscape}
\twocolumn

The above process led to the consolidation of five overarching modeling themes. In this section, I examine these main modeling themes, discussing the details and features of neural dialogue management models. Table \ref{tab:overview} maps each work to one or more modeling themes that are identified.

\subsection{Decoupling Semantics and Surface Realizations}

\label{sec:disentangling}
In this section, I review approaches that address an important limitations of vanilla end-to-end trained models of dialogue: separation of semantics and surface realization. In vanilla sequence-to-sequence modeling of dialogue, a trained neural network processes raw dialogue history as input and directly generates the subsequent natural language response without distinct intermediate steps for natural language understanding, state tracking, and generation. Crucially, traditional end-to-end training approaches fail to distinguish between "what to say" (semantics) and "how to say it" (surface realization), leading to complications as a single concept can be expressed in numerous ways.





\subsubsection{Conditioning Response Generation on Intermediate Symbolic Representations}
\citet{he-etal-2018-decoupling} propose an approach that separates a dialogue system's strategy from the natural language surface realization in a negotiation setting. They employ \textit{two} distinct sequence-to-sequence models to represent the dialogue (see Figure \ref{fig:he-negotiation}). The first model operates directly on the tokens in natural language utterances, while the second model works with coarse dialogue acts corresponding to the utterances. First, a parser is trained to map an utterance, $x_{t-1}$, to a coarse dialogue act $z_{t-1}$, which captures the high-level dialogue flow. A dialogue act consists of an intent and optional arguments. For instance, the utterance "I would like to pay \$125 for it. How does that sound?" is mapped to the dialogue act \texttt{propose(price=125)}. The dialogue act ontology includes \texttt{inform}, \texttt{inquire}, \texttt{propose}, \texttt{counter}, and \texttt{agree}. Both sequences are modeled using LSTMs \cite{hochreiter1997long}. Given $z_{t-1}$ and the dialogue history, $x_{<t}$, the dialogue manager predicts the coarse dialogue act of its response, $z_{t}$. Finally, a natural language response is generated based on $z_t$.

\archdiagram{rq2}{he-negotiation}{In \citet{he-etal-2018-decoupling}, the natural language parser generates dialogue acts, $z_{<t}$, corresponding to the utterances $x_{<t}$ respectively. The next response is generated in two steps: (1) generating the next dialogue act, $z_t$ based on the previous dialogue acts, $z_{<t}$ and 2) generating the natural language response based on both the dialogue history, $x_{<t}$ and the preceding dialogue act, $z_t$.}

\citet{santhanam_learning_2020} present an approach that generates an intermediate "plan" to condition the natural language response. They demonstrate this method on the \textsc{Persuasion For Good} dialogue corpus \cite{wang2019persuasion}, in which the agent persuades a listener by offering compelling incentives. The process consists of two steps: first, generating a pseudo-natural language plan, and second, conditioning the model on this plan to produce the final response. Figure \ref{fig:santhanam} illustrates an example dialogue context along with a system-generated intermediate plan. In response to B's final utterance, the model first generates a plan with an \textit{ask} and \textit{framing}, such as \texttt{PERFORM [provides [relief]]}. It then creates a natural language response, such as \textit{"Save the Children is an international non-governmental organization that promotes children's rights, provides relief, and supports children in developing countries"} (surface realization).

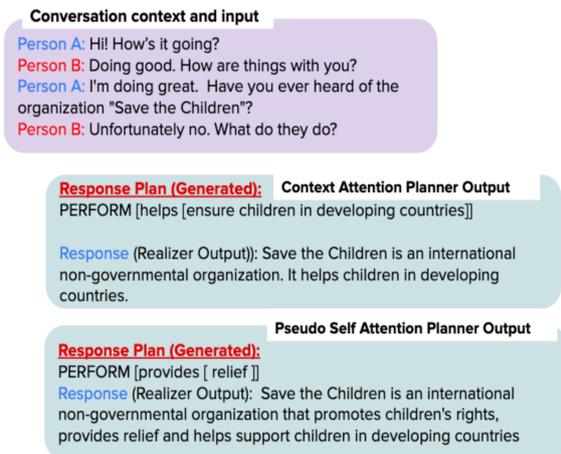
\archdiagramhalf[0.5]{rq2}{santhanam}{Separation of NLG into planning and surface realization \cite{santhanam_learning_2020}.}

\subsubsection{Conditioning Response Generation on Latent Representations}
In contrast to approaches that employ symbolic intermediate representations like dialogue acts and plans, some methods utilize latent variables instead.

\citet{Yarats2017HierarchicalTG} propose generating a short-term plan using latent sentence representations. The natural language response is then generated by conditioning the response generation step on this plan. To learn these latent sentence representations from data, the authors deviate from the traditional paradigm of maximizing the likelihood of the current utterance $x_t$ given the latent representation $z_t$. Rather, they train $z_t$ to maximize the likelihood of the response at the \textit{next turn}, $x_{t+1}$. This approach is motivated by the observation that while the traditional method effectively models semantic similarity, it fails to capture the influence of an utterance on the dialogue partner's response. This distinction can be crucial in negotiation settings, where even subtle lexical substitutions such as using "two" instead of "one," can significantly affect the final outcomes of the conversation.



\subsection{Incorporating Shared Artifacts, Environments and Knowledge}
Collaborative dialogues often take place when participants are working on a shared artifact or within a co-observed environment (e.g., an architect and a builder collaborating to build a Minecraft structure). Collaboration can also involve an asymmetry in the \textit{knowledge} that the participants are privy to (e.g., the architect may have a bird's eye view of the world, while the builder only has a first-person view). In this section, I provide a high-level overview of the approaches taken by collaborative dialogue systems to incorporate three common modalities: navigable environments, co-created artifacts, and text-based background knowledge\footnote{Note that I do not delve into fine-grained details of multimodal encoding and fusion methods. Several existing surveys on multimodal dialogue systems cover those topics, and I refer readers to works such as \citet{liu2022survey} for a comprehensive overview of such techniques.}. 

\subsubsection{Navigable Environments} \label{sec:navigable}
Visual-language navigation (VLN) \cite{park2023visual, lukin-etal-2018-scoutbot} is a broad research area that explores the interplay between natural language instructions and visual perception to enable autonomous agents or robots to navigate complex environments. Typically, a leader (e.g., a human user) provides natural language directions to a follower (e.g., a robot) navigating the environment. The follower may ask clarification questions to resolve ambiguities in either the environment or the instructions. In this section, I discuss two VLN systems, \citet{chi_just_2020} and \citet{de_vries_talk_2018}, focusing on their world and dialogue state representations, how these representations are used in dialogue management, and their approaches to handling clarification questions.

In \citet{chi_just_2020}, a human user instructs a navigating robot situated in a 3D environment with commands such as, \textit{``Walk straight, right before you reach the bed''}. The authors propose a navigating agent modeled as a Markov Decision Process (MDP) with states representing the agent's position and orientation. The agent's action space consists of environment-defined "navigable locations" and an additional clarification question action. Visual perception, position information, and user instructions serve as multimodal input to an LSTM model \cite{hochreiter1997long}, outputting a softmax distribution, $P^t$, over the action space. The authors explore both supervised and reinforcement learning methods for training.

In supervised learning, the environment calculates the shortest path $R_t = \{v_t, v_{t+1}, \dots, v_n\}$ from the current viewpoint $v_t$ to the target location $v_n$ at each time step $t$. Training loss is determined by the cross-entropy loss between the predicted action and the next action in the shortest path. For reinforcement learning, the agent selects its action by sampling from the action distribution $P^t$. A reward of +2 is granted if the final location is within 3 meters of $v_n$, and -2 otherwise. The Advantage Actor Critic (A2C) algorithm \cite{konda_actor-critic_1999} is utilized for RL training. The reward mechanism is further adjusted to encourage the agent to disambiguate its future trajectory by minimizing clarification questions.

A closely related work is that of \citet{de_vries_talk_2018}, where a Tourist agent and a Guide agent collaborate using natural language dialogue to reach a target location that is known only to the Guide. While the Guide has access to a 2D city map, the Tourist only has a first-person view of the local surroundings.


\subsubsection{Co-Created Artifacts}

Many collaborative activities involve co-creating some kind of artifact. Artifacts can range from block-based structures in a Minecraft game \cite{shi_learning_2022, narayan-chen_collaborative_2019} to drawings \cite{kim-etal-2019-codraw} to source code \cite{griffith_investigating_2022}. The two participants may be set up in either an asymmetric leader-follower setup or a symmetric one where both of them possess equal knowledge and skills.

Minecraft is a widely-used platform for studying collaborative dialogue in "Architect-Builder" scenarios, where the Architect agent instructs the Builder agent on constructing a target structure in the game environment. \citet{narayan-chen_collaborative_2019} introduce a dialogue model for the Architect that utilizes two types of counters—global and local block counters—to track the building progress. Global counters monitor overall block placements, upcoming placements, and removals, while local counters focus on a 3x3x3 region surrounding the Builder's last action. These block counter vectors, capturing attributes such as position and color, are processed through a fully-connected layer and combined with the word embedding vector for the decoder. This integration of dialogue history and world state representations enables the generation of contextually accurate instructions and a mental model of the target structure.

\citet{shi_learning_2022} explore generating clarification questions from the Builder's perspective. Unlike previous works, which prioritize Architect communication and Builder actions, their study addresses the generation of Builder utterances. The dialogue state consists of dialogue history, world state (with 3D voxels represented as 8-dimensional vectors), and the last action (encoded as a vector detailing action type, block color, and location). These three elements are transformed into vector representations through a convolutional neural network (CNN) and merged using a neural fusion module, resulting in the final dialogue state representation.

\archdiagramhalf[0.45]{rq2}{narayan-chen-2019}{Architecture Diagram from \citet{narayan-chen_collaborative_2019}.}

 \citet{kim-etal-2019-codraw} present a goal-driven collaborative task very similar to the Minecraft setting, involving two participants—a Teller and a Drawer—where the Teller describes a drawing to the Drawer, who then recreates it based on the instructions. The authors propose initial models for automating both agents, assuming the Drawer does not ask clarifying questions. The Teller's role involves understanding the drawing and generating natural language descriptions, while the Drawer is expected to comprehend the Teller's instructions and replicate the drawing. Neural models are proposed for both agents, with the Teller's model using a conditional language model with an LSTM and two attention modules that attend to individual clip art pieces in the drawing. The Teller model also incorporates an auxiliary training objective to track which parts of the scene have been communicated, using the LSTM's output at each utterance separator token to classify whether each clip art has been described or not.

\subsubsection{Textual Background Knowledge} \label{sec:text-background}
In many instances, it is common for the background knowledge pertaining to the collaborative task to be in the form of text. For example, in \citet{qiu_towards_2022}, a game agent is provided background and situational knowledge of the world using three pieces of text: the setting (e.g., \textit{The main foyer is massive. A grand staircase sits to the back of the foyer leading to the upstairs \dots}), the agent's self-persona (e.g., \textit{Servant. I come from the lower class. I do what I am told without question. I can not read. I have not seen my family in a long time. I carry a duster, a small bucket, a rag \dots}), the partner's persona (e.g., \textit{I am a king of the whole empire. I give rules and pursuit them. I am brave and fearless. I am carrying a crown and a scepter}).

The proposed model in \citet{qiu_towards_2022} encodes this text-based representation by feeding it into GPT-3 \cite{brown2020language}, a large language model (LLM) as part of its prompt. For example, the above description is fed into the model as follows: \textit{``The following is a conversation that happened in a game of Dungeons and Dragons: [Context] [DM Text] [Player Name]:[Player Ability Check] Question: What do you think that the DM is trying to guide the player to do by mentioning [Extracted Guiding Sentence]? Answer''}.

\subsection{Graph-Based Representations of Dialogue State} \label{sec:rq2-graph}
Graph-based dialogue state representations strive to strike a balance between rigid data structures, such as slot-value pairs, and completely unstructured continuous vector representations. By employing graphs, these representations closely mirror the inherent structure of real-world tasks or domains. This approach offers semi-interpretable and robust neural representations, blending the advantages of visualizations with the potential for adaptable representation. Such representations learn entity representations from data, and their popularity has surged in recent years for task-oriented dialogue applications \cite{walker_graphwoz_2022, andreas_task-oriented_2020}. In this section, I discuss the components (nodes and edges) of a graph and the methods employed to update the graph as the dialogue unfolds.

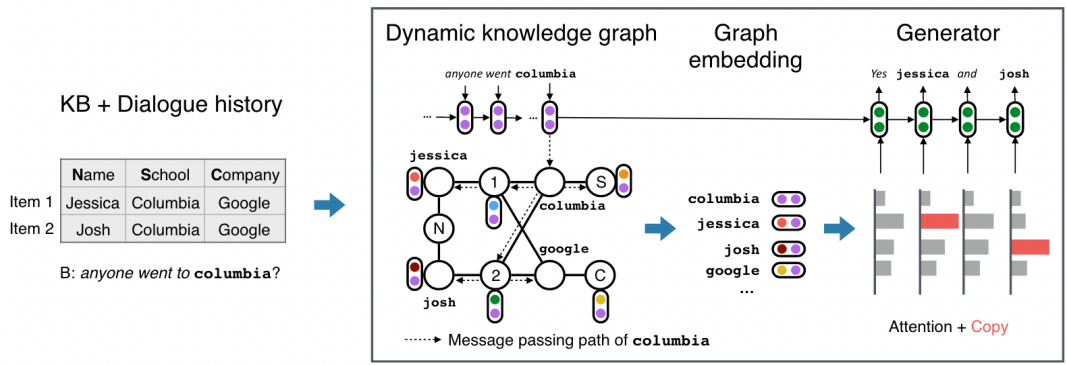
\archdiagram[0.9]{rq2}{he-2017}{Architecture of symmetric collaborative dialogue state from \citet{he_learning_2017}.}

\subsubsection{Graph Representation}
Among the reviewed works, \citet{he_learning_2017} uses a graph-based dialogue state representation for a task that involves identifying a mutual friend from a knowledge base through text-based dialogue. Their proposed approach consists of a knowledge graph, graph embeddings, and an utterance generator. The knowledge graph comprises the following three types of nodes, with edges between them representing their relations. When an utterance $t$ mentions a new entity, it is added as a new node to the knowledge graph. 
\begin{enumerate}
    \item entity nodes representing people, universities, and companies,
    \item item nodes corresponding to a single row-entry in the agent's knowledge base (e.g., a friend), and
    \item attribute nodes that represent relationships between entities, such as a person's connection to a university
\end{enumerate}

In \citet{qiu_towards_2022}, a knowledge graph is used to represent the agent's mental state at a given point in a game. The graph is initialized by parsing a text-based description of the agent's persona into constituent entities and relations using a rule-based parser. The nodes in the graph correspond to the agents involved in the game, persona descriptions, objects, descriptions of the objects, and descriptions of the environment. The edges describe the relationships between the agents and the objects as present in the agent's mental state. Figure \ref{fig:qiu-2022-mental} shows an example ``mental state graph.''

\subsubsection{Graph Encoding and Updating}
In \citet{he_learning_2017}, the final dialogue state representation is obtained from the dialogue state graph's node embeddings (Figure \ref{fig:he-2017} provides a sample snapshot of the dialogue state used in this system). Given a dialogue of $T$ utterances, $T$ graphs $(G_t)_{t=1}^{T}$ are constructed for each agent $A$. The graph $G_t$ is constructed by updating $G_{t-1}$ with any new entities not previously mentioned. To obtain a latent representation of the graph, an embedding is associated with each node in the graph, $v \in G_t$. A node embedding $V_t(v)$ is built from structural properties of an entity in the knowledge base, embeddings of the utterances in the text-based dialogue history, and the message passing between the graph nodes. Each of these is elaborated below.
\begin{enumerate}
    \item \textbf{Structural properties:} The structural properties of a node $v$ are encoded using a featurization function $F_t(v)$, which encodes various pieces of information such as the degree, type and number of mentions of the node.
    \item \textbf{Mention vectors:} A mention vector of a node $v$ contains an aggregate representation of the mentions of that entity so far in the dialogue history. At every dialogue turn $t$, the mention vector $M_t(v)$ is set to $M_{t-1}(v)$ if utterance $t$ does not contain mention of the entity and a soft weighted sum of $M_{t-1}(v)$ and $\tilde{u}$ if the utterance does contain mention of the entity. $\tilde{u}$ is obtained by using an LSTM to encode the utterance.
    \item \textbf{Recursive node embeddings:} It is important to take into account information from neighboring nodes in the knowledge graph. For example, to answer the question \textit{``anyone went to colombia?''} as in Figure \ref{fig:he-2017}, the model should take cognizance of the entity nodes of type \textit{person} who went to \textsc{colombia} university (in this case, \textsc{jessica} and \textsc{josh}). This is achieved through a mechanism known as belief propagation.
\end{enumerate}

In \citet{qiu_towards_2022}, the graph update step is broken down into two update types: discrete and continuous. The actions in the game environment, which can be converted into simple, text-based templates are modeled as discrete graph updates, while the unconstrained natural language utterances, which are harder to model, are incorporated to the representation through a continuous graph update instead.
\begin{enumerate}
    \item \textbf{Discrete update:} In the discrete update step, $\delta g_t$ is a set of ADD and DEL operations. Each action in the game environment is parsed into a set of these atomic operations over the graph. For example, the action \textit{``give object to agent''} is parsed into DEL($actor, object, carrying$) and ADD($agent, object, carrying$).
    \item \textbf{Continuous update:} The continuous update step is motivated by the reasoning that an utterance can also have implicit effects on the agent's mental state. This step is handled by a recurrent neural network in the following steps:
    \begin{align*}
        \delta g_t = f_{\delta}(h_{G_{t-1}}, h_{O_t}) \\
        h_t = \text{RNN}(\delta g_t, h_{t-1}) \\
        G_t = \text{MLP}(h_t)
    \end{align*}
    Here, $f_{\delta}$ is a bi-directional attention layer \cite{yu2018qanet} that aggregates information from the previous graph state and the latest set of observations to generate the graph update, $g_{\delta}$. A graph convolutional network (GCN) \cite{schlichtkrull2018modeling} and a BERT-based text encoder is used to encode $G_{t-1}$ and the text-based dialogue history into their latent representations $h_{G_{t-1}}$ and $h_{t-1}$ respectively. The RNN is modeled by an LSTM \cite{hochreiter1997long}, which models an intermediate hidden state, which is finally used by a multi-layer perceptron (MLP) to generate the updated graph.
\end{enumerate}


\subsection{Incorporating Domain-Specific or Expert Knowledge}
Neural dialogue management methods aim to be data-driven without manual feature engineering. Nonetheless, encoding domain-specific or expert knowledge as inductive biases can be advantageous, as it may accelerate learning or reduce data dependency. In this section, I examine various approaches taken in the selected works to incorporate domain-specific or expert knowledge into neural dialogue managers.

\subsubsection{Utterance-Level Labels}
One straightforward method to encode domain knowledge into neural dialogue managers is through the use of handcrafted taxonomies for utterance classification. For instance, in a negotiation context, \citet{zhou-etal-2019-dynamic} initially classify each dialogue utterance into one or more \textit{tactics} from a predefined set. This set of tactics, derived manually from behavioral science research, includes tactics such as \textit{communicate politely}, \textit{use hedge words}, and \textit{show dominance}. A sequence model is trained on the sequence of these tactic labels to determine the subsequent dialogue move. Similarly, \citet{zhou_augmenting_2020} transform utterances into a sequence of tactic and strategy labels for another negotiation scenario.

\subsubsection{Vector Featurization}
Domain knowledge can also come into play while designing one-hot representations preceding an embedding layer. \citet{kim-etal-2019-codraw} develop a neural dialogue model to act as a \textit{Teller} agent that describes a clip-art drawing to a \textit{Drawer} through a natural language dialogue. From human-human dialogues in the same environment, they observed that humans use compositional language to describe clip-art configurations and attributes (e.g., \textit{"where is trunk of second tree, low or high"}). To help understand such utterances, each clip-art piece in the scene is represented using a vector that is a sum of learned embeddings for attributes such as \textit{type=Mike}, \textit{size=small}, etc.

Another example of domain-specific vector featurization is in developing an Architect for the visual-language navigation setting described by \citet{shi_learning_2022}. Each voxel in the 3D Minecraft environment is first encoded using a one-hot vector that captures the voxel's 3D position, its colors, and its neighboring voxels. Along with that, their representation also includes the location at which the Builder last made a move. This is motivated by the observation that Builders often refer to their recent actions in their utterances.

\subsubsection{Response Ranking}
\citet{qiu_towards_2022} train a dialogue agent to behave in a ``socially intelligent'' way. Their agent is situated within a large-scale, crowd-sourced fantasy text-adventure game called LIGHT \cite{urbanek-etal-2019-learning}. In their work on intercultural research, \citet{schwartz1992universals} identify a set of universal basic human values that transcends cultures (see Figure \ref{fig:schwartz-human-values}), which \citet{qiu_towards_2022} adopt in the response generation step of their model. Specifically, a learned \textit{value function} $f_v(\cdot)$ ($v \in \{achievement, power, security, \dots\})$ takes in an action (or utterance) and outputs a scalar value. The value function is learned from a knowledge base of human values from the \textsc{ValueNet} dataset \cite{qiu2022valuenet}. During response generation, the agent first narrows down the action space based on feasibility, and then ranks the filtered actions based on the equation below (here, $p$ is a text-based description of the agent's persona that the agent is seeded with (elaborated in Section \ref{sec:text-background})). The main purpose of doing the second step is to ensure that the agent's responses are consistent with its own value system.
\begin{equation*}
    u(a_i) = \sqrt{\sum_{v \in V}{(f_v(p) - f_v(a_i))^2}}
\end{equation*}

\archdiagramhalf[0.4]{rq2}{schwartz-human-values}{Theory of basic human values used in \citet{qiu_towards_2022}'s socially intelligent agent.}

\subsubsection{Reward Functions for RL}
Another way to incorporate domain-specific knowledge is through the use of handcrafted reward functions for reinforcement learning. In Cuayahuitl et al. (2015), a dialogue agent is trained to engage in conversations within the context of the Settlers of Catan game, focusing on resource trading. An example trading offer might be, "I will give anyone sheep for clay." The reward function is manually defined within the environment through a user simulator. Specifically, the reward, $r$, is constructed as a piecewise function that depends on the number of points obtained in the game. This information is used to train a deep reinforcement learning model.

\begin{equation*}
r =
\begin{cases}
GainedPts \times w_{gp}, & \text{if } GainedPts > 0 \\
TotalPts \times w_{tp}, & \text{otherwise}
\end{cases}
\end{equation*}

In the reward function above, $w_{gp}$ and $w_{tp}$ represent weighted hyperparameters that are manually set by the authors. The function assigns a reward based on the number of points gained during a specific action (GainedPts) or the total points accumulated in the game (TotalPts), depending on whether any points were gained during that action.

\subsection{Theory-of-Mind Modeling} \label{sec:rq2-tom}
Theory-of-Mind (ToM) \cite{premack1978does} is a concept in cognitive science that refers to the ability to attribute mental states such as beliefs, desires, intentions, and emotions to oneself and others, understanding that these mental states may differ among individuals. The incorporation of a Theory-of-Mind model into a collaborative dialogue manager is closely related to the BDI (beliefs, desires, and intentions) framework  \cite{gratch2000emile} used in plan-based approaches. 

This section examines the chosen works to detail how they incorporate ToM in their dialogue management. Specifically, it discusses two things: 1) the aspects of the mental state captured and 2) the methods used to collect labels for training machine learning models that can predict theory-of-mind variables from the dialogue context.

\subsubsection{Dimensions of Mental State}
Among the reviewed works, the aspects of mental states that are modeled include both task-related (e.g., what a participant is currently working on or believes the task state to be) as well as aspects related to a participant's personality and emotion \cite{qiu_towards_2022}.

In \citet{bara_mindcraft_2021}, the authors investigate a setting involving two partners (Player A and Player B) collaborating to create a target material in a Minecraft environment by mining and combining blocks. Owing to the asymmetry in both their skills and knowledge, the participants engage in rich natural language dialogue to develop a strong mental model of the task requirements, their own task-state, their partner's knowledge, and their partner's task-state. Specifically, the proposed ToM model in their paper seeks to capture 1) a player's own task status (whether a specific material has already been created or not), 2) a player's self-knowledge (whether a player knows how to create a material or not), and 3) their partner's current task (what a player thinks their partner is working on at a given point in time).

The setting of dialogue in \citet{qiu_towards_2022} is a fantasy text-adventure game, LIGHT \cite{urbanek-etal-2019-learning}, in which agents can talk to other agents in free-form text, take a closed set of actions, or express certain emotions. In this setting, \citet{urbanek-etal-2019-learning} present a ``socially-aware'' response generation model that explicitly models the mental state of the agent as a time-varying graph. The graph is constructed (and updated) based on three sources: 1) an initial persona description associated with the agent, 2) the dialogue history, and 3) the game state. A sample graph from their model is shown in Figure \ref{fig:qiu-2022-mental}. The nodes correspond to natural language descriptions of agents and objects in the world.

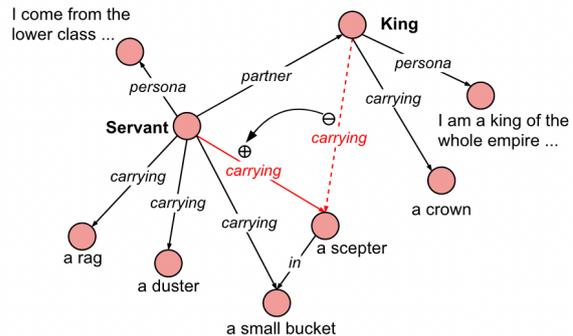
\archdiagramhalf[0.50]{rq2}{qiu-2022-mental}{Graph-based mental state representation of the Servant agent in the LIGHT fantasy text-adventure game (from \citet{qiu_towards_2022}).}

\citet{zhou_ai_2022} develop an agent that can play the role of a Dungeon Master (DM) in a Dungeons-and-Dragons (D\&D) game. The agent's goal is to produce utterances such that the remaining players are incentivized to perform a specific action that the DM has in mind. This action, in turn, is designed to make progress in the story that the DM has constructed. Therefore, the specific aspect of the player's mental model that is modeled by the DM agent in this work is the following: \textit{How will the player react to my utterance?}

\subsubsection{Label Collection Methods for ToM Modeling}
Collecting training labels is a crucial challenge when training machine learning models for ToM modeling. In this section, I discuss three approaches identified from the reviewed works.

\paragraph{Simulated Environment.} In the agent proposed by \citet{qiu_towards_2022}, situated within a simulated game, the dialogue agent has full visibility into the game state. This visibility enables the construction and updating of the mental state representation directly from the available information. The agent's mental state is represented as a graph, initialized by parsing the "seed" natural language description of the agent's persona. As the conversation progresses, the dialogue history and game state are used to incrementally update this graph representation.

\paragraph{Human-in-the-Loop.} In the setting proposed by \citet{bara_mindcraft_2021}, dialogue agents operate within a Minecraft world and collaborate with a partner to combine primitive blocks into a target block. To collect labels for ToM modeling, they employ an online human-in-the-loop data collection methodology, prompting users to reveal their mental models during the interaction. A machine learning model is then trained to predict the mental models of both partners based on dialogue and task context. The proposed model utilizes either an LSTM or a Transformer to process inputs from three different modalities: a video feed encoded with a convolutional neural network, knowledge in the form of a graph encoded using a gated recurrent unit (GRU) network \cite{chung}, and dialogue utterances encoded with a BERT model. User-prompted questions at 90-second intervals are also encoded and combined with the Transformer representation, which is then fed into a feed-forward network predicting the partner's ToM attributes. This architecture is depicted in Figure \ref{fig:bara-2021-1}.

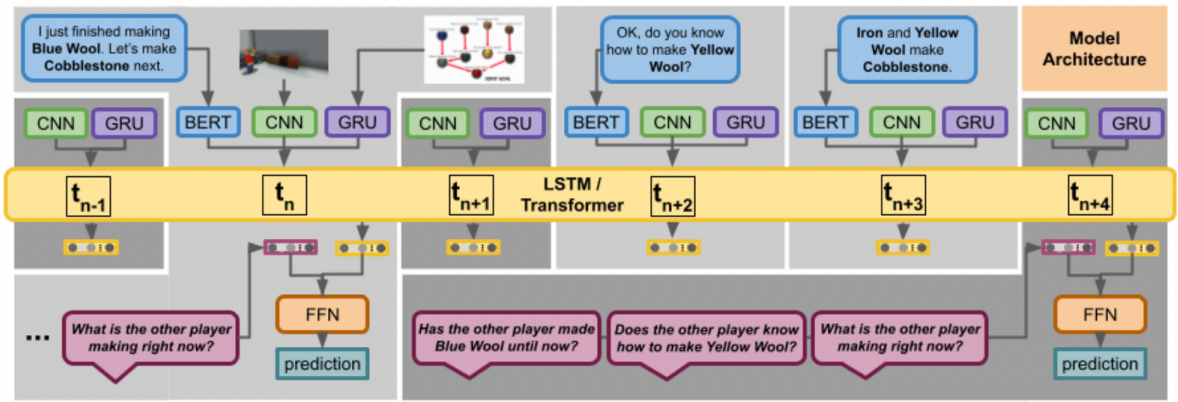
\archdiagram{rq2}{bara-2021-1}{Architecture of the ToM-prediction model proposed in \citet{bara_mindcraft_2021}.}

\paragraph{Leveraging LLMs to Generate Labels.} \citet{zhou_ai_2022} propose a method designed to circumvent the expensive nature of human-in-the-loop approaches. The dialogue agent plays the role of a Dungeon Master (DM) in a Dungeons and Dragons game, with the objective of generating utterances so as to generate utterances that maximize the likelihood of players making a specific move intended by the DM. The agent predicts the players' next moves to a given utterance using a set of language models working together to generate affordable proxies for user responses. The process consists of the following steps:
\begin{enumerate}
    \item Annotators label a small number of Dungeon Master's (DM's) utterances in a collected corpus of interactions, identifying the specific span that influenced the players' actions.
    \item An "inverse dynamics model (IDM)" is trained over these labels to predict the specific span that influenced the player's actions, given their next action (from the future).
    \item An intent generator is trained by first mining intents from the DM's utterances and then training another model over the mined intents.
    \item A player model is trained to predict the player's next move based on the dialogue history.
    \item Lastly, the DM model is trained using reinforcement learning (RL) to generate utterances that lead to the player performing an action aligned with the DM's intent.
\end{enumerate}

Both the IDM and the intent generator are trained by prompting GPT-3 \cite{brown2020language} to 1) output excerpts from the DM's utterance and 2) generate natural language descriptions of intents, respectively.

\section{Discussion}
\label{sec:discussion}

This section summarizes the findings from the literature review and discusses open challenges and future research directions for neural collaborative dialogue management.

\subsection{General Trends}

I begin by outlining some general trends uncovered through this literature review.

\paragraph{Historical Trends.} Most of the works that laid the foundation for collaborative dialogue systems (outlined in Section \ref{sec:rq1}) came about in the period from the late-1980s to the mid-2000s. Throughout the late-2000s and the 2010s, dialogue systems research became predominantly focused on slot-filling assistants and open-domain chatbots, resulting in limited advancements in both theoretical and practical frameworks for collaborative agents. Similar observations about the research trends have been echoed by notable researchers in the field \cite{cohen_foundations_2019, papaioannou2018human}.

\paragraph{Amount of Research Activity.} The search for works proposing neural network-based techniques in constructing collaborative agents yielded only a small number of results, despite the flurry in research activity for neural network based techniques for task-oriented and open-domain dialogue systems during the same time period. I hypothesize three reasons for this disparity:
\begin{enumerate}
    \item The lack of straightforward automated evaluation methods in complex collaborative dialogue settings (elaborated below in Section \ref{sec:discussion-eval})
    \item Immediate commercial impetus for task-oriented and open-domain dialogue provided by the proliferation of ``virtual assistants'' such as Amazon Alexa, Apple Siri, and Google Home, and
    \item The lack of collaborative analogues of large benchmark corpora such as MultiWOZ \cite{budzianowski2018multiwoz} and Schema-Guided Dialog \cite{rastogi_towards_2020, mehri_schema-guided_2021} as well as shared community challenges such as DSTC \cite{williams_dialog_2014}, which were instrumental in catapulting research activity in task-oriented dialogue systems.
\end{enumerate}

\subsection{Challenges to Holistic Evaluation} \label{sec:discussion-eval}

We see that in all the works identified under RQ 2, the tasks, domains, environments, and objectives are carefully curated so as to facilitate automated, quantitative evaluation. However, in this process, the authors are often forced to make simplifying assumptions that might not be reflective of the full complexity of the real-world. For example, a common simplifying assumption made in Leader-Follower setups such as \cite{kim-etal-2019-codraw, chi_just_2020, narayan-chen_collaborative_2019, narayan-chen_towards_2017} is that the Follower cannot initiate dialogue, but can instead only receive instructions, or at best, ask clarification questions.

As another example, \citet{sanders_towards_2022} observe, in the context of a persuasion dialogue for charity donation, that a strict definition of success such as the amount of money donated by the latter may not accurately capture the performance of the Persuader agent since the Persuadee's decision can be influenced by factors out of the Persuader's control, such as the Persuadee not having money to donate. In such a situation, holistic, automated evaluation can be challenging. Such difficulties in evaluation is a general trend among most collaborative dialogue settings excluding a few well-defined cases such as certain types of negotiations, where the outcomes can be captured quantitatively by the end result of the dialogue.

In general, the roadblock for holistic evaluation is that on one hand, automated measures can only evaluate individual modules of a dialogue system or individual aspects of what makes a successful collaborative dialogue. On the other hand, complete end-to-end human evaluation is expensive. Existing works mostly take the former, piecemeal approach to evaluating a dialogue system, which, albeit useful, may not be fully capture the features that stakeholders ultimately care about.


\subsection{Combining the Strengths of Neural and Plan-Based Approaches}
The works identified for RQ 2 reveal a clear trend of incorporating features typically found in plan-based approaches to dialogue management. Specifically, the modeling approaches proposed in these works, such as decoupling semantics from surface-level representations (Section \ref{sec:disentangling}), moving towards human-interpretable representations of dialogue state (Section \ref{sec:rq2-graph}), and incorporating Theory-of-Mind models (Section \ref{sec:rq2-tom}), can all be mapped one-to-one with constructs that are ``natively available'' in plan-based dialogue managers (Section \ref{sec:rq1-planning}). Plan-based dialogue management techniques (Section \ref{sec:rq1-planning}) employ discrete action-spaces, can explain their decisions \cite{shams, nothdurft, honold, nothdurft-etal-2015-interplay}, and maintain explicit notions of their partners' beliefs through the use of belief-desire-intention (BDI) models in their architectures \cite{cohen_planning-based_2023}. Given this inherent applicability of plan-based dialogue management for collaborative dialogue, exploring ways to combine the strengths of both approaches presents a promising area for further research. Neural approaches can learn directly from data and provide high representational power and flexibility, while plan-based approaches are recognized for their reliability, consistency, and dependability.




\subsection{LLMs for Data Generation}

\archdiagram{generic}{chat-gpt-labeling}{\citet{gilardi2023chatgpt} shows that ChatGPT outperforms M-Turk crowdworkers when it comes to inter-annotator agreement against a trained annotator.}

Deep learning has achieved remarkable success in various NLP tasks, including question-answering, text classification, and natural language generation. Nevertheless, dialogue management remains an unresolved problem, as collecting extensive data for new domains is often exceptionally difficult and impractical. Although reinforcement learning-based approaches can learn from interactions without a large corpus to begin with, deploying sub-optimal dialogue policies in sensitive areas such as medicine, aviation, and education may lead to detrimental or even fatal consequences. However, the potential of large language models (LLMs) could transform this situation.

Traditionally, gathering training data for neural dialogue models in specific domains involves either using a small number of in-house researchers as annotators (e.g., graduate students) or enlisting larger groups of individuals through crowdsourcing platforms like Amazon M-Turk. Both methods demand substantial effort and incur significant costs. This issue becomes even more pronounced when deep domain expertise is required for annotation, as crowdsourced annotations from the general populace may not be reliable. Recently, \citet{gilardi2023chatgpt} demonstrated that instruction-tuned LLMs, such as ChatGPT, outperform crowdworkers in annotation quality for several NLP tasks. Figure \ref{fig:chat-gpt-labeling} compares ChatGPT and M-Turk crowdworkers in terms of inter-annotator agreement with trained annotators. Advancements in harnessing large language models for annotating data, creating new datasets, and augmenting existing ones are promising directions to help address this challenge.

\subsection{Absence of a Software Ecosystem}
Finally, despite over four decades of research, the software landscape for practical authoring tools and development frameworks for developing collaborative dialogue systems is still largely limited to research prototypes and narrow domain-specific techniques. Developing general-purpose tools for system designers to quickly prototype an agent in a completely new domain continues to be non-trivial even today. This situation is in striking contrast to the current landscape for form-filling and open-domain assistants, where widely accessible commercial tools such as Google DialogFlow\footnote{\href{https://cloud.google.com/dialogflow}{https://cloud.google.com/dialogflow}}, Amazon Lex\footnote{\href{https://aws.amazon.com/lex/}{https://aws.amazon.com/lex/}}, Rasa\footnote{\href{https://rasa.com/}{https://rasa.com/}}, LUIS\footnote{\href{https://www.luis.ai/}{https://www.luis.ai/}}, and ParlAI\footnote{\href{https://parl.ai/}{https://parl.ai/}} empower individuals without expertise in dialogue systems or natural language processing to create agents suitable for a vast range of scenarios.

Schema-driven dialogue system design \cite{kane_flexible_2022, mehri_schema-guided_2021, rastogi_towards_2020} and prompt-based techniques \cite{kaplan_in_context, wei2022emergent} represent two promising classes of interfaces for dialogue system authoring. However, to determine the optimal type of interface for a specific user and domain, further human-centered research is needed, including user studies that encompass diverse user groups and various dialogue systems.





\section{Summary}
\label{sec:conclusion}
This paper begins by reviewing foundational paradigms for developing collaborative dialogue agents that have emerged over the past four decades of dialogue systems research. Each dialogue paradigm is described, and the applications in handling collaborative dialogues are illustrated. The focus then shifts to data-driven, neural approaches for collaborative dialogue management that require minimal handcrafting. The limitations of "vanilla" end-to-end neural networks and slot-based methods for dialogue state representation are discussed, serving as motivation for reviewing neural modeling approaches for collaborative dialogue. From a selection of \numworks works, this review identifies five core modeling themes and examines the implementation of each theme by the respective studies in specific domains. Lastly, the overarching trends are analyzed, and the paper concludes by presenting a set of open challenges and future research directions for the field of collaborative dialogue management.



\section*{Acknowledgement}
I would like to thank Dr. Kristy Elizabeth Boyer and Dr. Bonnie Dorr for their invaluable advice. I would like to also thank Dr. Anthony Botelho for his helpful inputs during the initial stages of organizing this literature review.


\bibliography{references2, custom}
\bibliographystyle{acl_natbib}

\end{document}